# Autonomous Aggregate Sorting in Construction and Mining via Computer Vision-Aided Robotic Arm Systems

Md. Taherul Islam Shawon [a,1], Yuan LI [a,1], Yincai CAI [a,1], Junjie NIU [a], Ting PENG [a,*]

[a] *Key Laboratory for Special Area Highway Engineering of Ministry of Education, Chang'an University, Xi'an, 710064, China*

A B S T R A C T

Traditional aggregate sorting methods, whether manual or mechanical, often suffer from low precision, limited flexibility, and poor adaptability to diverse material properties such as size, shape, and lithology. To address these limitations, this study presents a computer vision-aided robotic arm system designed for autonomous aggregate sorting in construction and mining applications. The system integrates a six-degree-of-freedom robotic arm (Hiwonder JetArm), a binocular stereo camera (Orbbec Gemini) for 3D perception, and a ROS-based control framework. Core techniques include an attention-augmented YOLOv8 model for aggregate detection, stereo matching for 3D localization, Denavit–Hartenberg kinematic modeling for arm motion control, minimum enclosing rectangle analysis for size estimation, and hand–eye calibration for precise coordinate alignment. Experimental validation with four aggregate types (limestone, marble, sandstone, granite) achieved an average grasping and sorting success rate of 97.5%, with comparable classification accuracy. Remaining challenges include the reliable handling of small aggregates and texture-related misclassification. Overall, the proposed system demonstrates significant potential to enhance productivity, reduce operational costs, and improve safety in aggregate handling, while providing a scalable framework for advancing smart automation in construction, mining, and recycling industries.

## 1. Introduction

The blistering pace of development of automation technologies and in specific the technologies of robotics and computer vision fundamentally altered the way industrial operation in general should take place, as it allows previously unfathomed levels of efficiency, precision, and even flexibility in an industrial operation. One of the most influential trends is the implementation of computer vision-powered robotic arm systems to sort aggregates via a process that is core to the operations of construction, mining, recycling, and manufacturing industries because the quality of products and the effectiveness of their production and flow [1], directly depend on the speed and precision of their sorting [2]. Design of these systems and a little experimental work is an example of a multidisciplinary design with contributions in mechanical design, electronics, computer science, artificial intelligence, and control systems all playing a role in solving the challenging problem which is the real-world sorting problem.

In their conventional sense, aggregate sorting has so far depended exclusively on manual labor or incompetent mechanical systems that do not have the versatility and the intelligence to adapt to the different kinds of materials and slice unpredictable input streams. Manual sorting is tedious, inaccurate, and workers are subjected to dangerous conditions and conventional automatic systems can be restricted by the fixed rules, rigid equipment and they do not deal with the never-ending nature of aggregates e.g. different colors, textures, or odd shapes. The introduction of the computer vision-based robotic arms is a remarkable game-changer because it utilizes innovative imaging and intelligent algorithms to see, understand, and control objects in a degree of versatility and dexterity that have never been possible before.

The key to a computer vision-based robotic arm system is the enhanced partnership between visual awareness and robot manipulator. The basic structure of the system generally consists of an arm, and sometimes also end-effectors (e.g. grippers or suction cups), a vision module (can contain cameras, sensors and lights), and a computational module that performs image processing, decision-making and motion planning [3-7]. The vision module captures video stream or real-time images of the aggregate materials being heavy on a conveyor device or they are being presented in an area of work. These inputs are input into sophisticated image processing algorithms anything as simple as contour detection and thresholding all the way up to object recognition via deep learning [8-10]. This information is then processed by the computational unit to find, assign and locate each aggregate so as to know the best points of grasping and where to sort it.

Some of the important considerations in the design of such system include. Technically, the robotic arm needs to be able to accommodate a wide range of various sized and weighted aggregate with enough degrees of freedom, reach, and ability to carry a payload and be repeatable enough. The end-effector design should have the capability of safe and non-destructive manipulation especially when manipulating fragile or oddly shaped material. The system will have to have good power management, real-time communication and sensor integration electrically. Software wise, computer vision and computer control algorithms require effective calibration, synchronization, and error correction in order to support high accuracy and throughput conditions of different lighting constraints, occlusion, and back-ground clutter.

---

\* Corresponding author at: Key Laboratory for Special Area Highway Engineering of Ministry of Education, Chang'an University, Xi'an, 710064, China

E-mail address: t.peng@ieee.org (Ting PENG).

[1] Indicate equal contribution.

An experimental study of the assembly and testing of a computer vision-based robotic handle that can be used to sort aggregates requires a stringent approach which includes the simulation, prototyping, and testing. The first simulations can enable the optimization of the kinematic parameters, workspaces configuration and the behaviour of algorithms in deterministic conditions. The physical prototyping includes the task of connecting the robotic arm, continuation of the vision system, and sketches of the control software. Representative aggregate samples are then used to conduct experimental trials and important performance indices like sorting accuracy, cycle time, reliability and adaptability are statistically evaluated. These tests do not only confirm the functional requirements, but also reveal real problems, and real-world difficulties, e.g. sensor noise, mechanical wear, or unforeseen behaviors of the materials, that can be incorporated in the next round of design improvements.

Such automatic systems have much more of a reach. Computer vision-guided robotic arms boost productivity, minimize the costs of operations, and increase workplace safety by allowing high-speed, high- precision sorts that require a small human input. Application in dimensions where aggregate quality is the most important, including concrete manufacture where the coherence of the sand, gravel, and broken stone influences the structural integrity, automated sorting ensures that the quality is always up to the rigorous standards and there is less wastage of the material. Waste management and recycling Second, when it comes to waste management and recycling, intelligent sorting will help to separate valuable resources from mixed waste streams efficiently, promoting sustainability agendas and resource recovery practices [8, 10]. Moreover, these systems are modular and programmable to enable fast re-configuration to support new kinds of aggregates, sorting criteria or production needs, and offer the degree of flexibility, which is critical to agile works of modern manufacturing facilities.

Artificial intelligence and machine learning also help to enhance the functions of a computer vision-controlled robotic arm. Object detection and classification can be strongly robust (under noise, occlusion, or overlapping materials) with trained deep lean models on large datasets of aggregate images. The robotic arm projected on screen uses reinforcement learning and adaptive control algorithms so that the robotic arm can learn through experience to improve its performance in terms of grasping and sorting. The integration of stereo vision or 3D imaging technology lets to accurately estimate depth and perform spatial reasoning so that the system could be applied to complex scenes containing several partially invisible objects [10, 11]. These developments do not only enhance the accuracy and reliability of sorting, but also introduce new avenues of smart decision making, preventive maintenance, and automatic system adjustment.

Notwithstanding these developments, there are still a number of issues when it comes to designing and implementing a computer vision-directed robotic arm system to sort aggregate. The irregularity and uncertainty of the form of the aggregates present in the real world, the dust and fine particles to large irregular rocks, require sturdy strategies of sense and manipulation. The difference in lighting, dust and other environmental factors can distort the quality of images thus embedding advanced preprocessing and sensor fusion algorithms. Industrial sorting has real-time demands which place strong demands on system latency and computational efficiency necessitating hardware acceleration of optimised algorithms. System design is further complicated by the integration with other production lines and adherence with industrial standards, and safety regulations.

These challenges are met by continuous research and development activities in a few areas. Of utmost interest is the development of advanced vision algorithms, which can deal with wide range of material properties and environments. The versatility of handling is augmented by future design of adaptive end-effectors, end-effectors that are dynamically adjusted to use grip force and configuration in response to object design. Scalability and robustness of the implementation are enhanced by the use of distributed control architecture and hierarchical communication protocols i.e. master-slave systems using embedded microcontrollers [8, 10]. Experimental research is of importance in rating the system performance, failure mechanisms, and the performance validation of proposed solutions under the real tests of operation.

The algorithm and experiment in a computer vision-guided robotic arm assembly to conduct aggregate sorting is the embodiment of the potential innovation that smart automation can create in the industrial sector [12]. These systems do solve the inadequacies of traditional sorting techniques and comprise an effective blending of perception, cognition, and action that enables them to be highly accurate, fast and adaptive after surmounting the shortcomings of traditional sorting techniques. The fact that the challenge is multidisciplinary in nature brings about innovation in the interface of robotics, computer vision, artificial intelligence and systems engineering which are now moving towards introducing fully autonomous and intelligent manufacturing environments. As the technical and practical issues are being worked out, the popularization of such systems is expected to transform the sorting process in industry, increase the efficiency of raw materials use, and make operation in a more safe, sustainable way possible in a vast range of sectors of commerce [8-10].

## 2. Aggregate Image Acquisition

### 2.1. Aggregate Selection

The selection criteria are needed to be established before choosing the aggregates. Our points of criteria predominantly have two. The aggregates chosen should be representative in the first place. Nature has a variety of rocks hence not all these types of rocks can be used in experiments. Thus, we select typical rocks, which are very much utilized in engineering. Second, the aggregates are relatively differentiated in nature. Because target detection involves obtaining the characteristics of the target object to be classified, then there would be a problem in accuracy of detection in the case when there are no obvious features or the features are relatively chaotic. Against the above standards, every type of aggregate that is chosen in this paper is a rock which also happens to be frequently applied in road engineering like granite, sandstone, marble, and limestone. All these rocks have their characteristics concerning physical and chemical properties, apply widely within road engineering and their features are good as illustrated in Fig. 1.

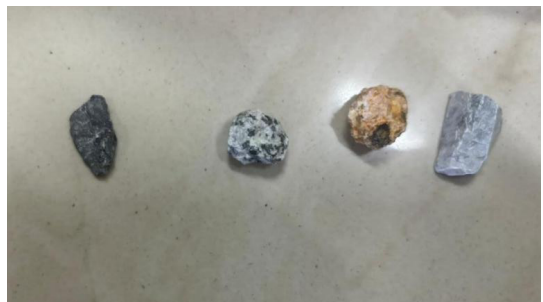

Fig. 1. Aggregate Selection

They form limestone, granite, sandstone and marble, left to right in the image.

Particle size of the chosen aggregate should also be determined after a determination of the type of aggregate. In order to be applicable in the field of practical engineering as well as the further grasping of the aggregate by the robotic arm, the particles size of the aggregate ought to be selected somewhere between 1 and 4 cm. The aggregates are then divided into three grades based on the particle size; one grade that has the particles of 1 cm and above, another 2 cm and above and the third is 3 cm and above, in that order and given numbers 1, 2 and 3. This allows it to provide comparative data to facilitate the accurate analysis (measuring) of the aggregate points size in any further experiment. Here in this case nearly one thousand aggregates have been gathered, and this is all that has been required in the subsequent experiments.

### 2.2. Creating a Dataset

Photos are saved when doing the capturing with names such as WIN_20240701_15_49_58_Pro; this shows that it is windows, the date, time the photo was taken, the order and the type of camera. But with the amount of photos that are likely to be taken being really large, this way of naming the photos will result in not being really able to tell where and with which kind of material they are taken.

Thus, there is need to rename the images. In this case, we will be able to batch rename through codes. The rename () function of the os library has been used to renaming in batch mode in this article. Among them, D is the type of rock, which is a marble, D-1 means the particle size of 1cm as the type of marble, and the last one is a sequence of the image. Granite is again symbolized as H, sandstone as S and Limestone as SH. With this naming convention all the 1219 images will be renamed so that the later required labelling files will easily be made.

LabelImg is a Python based open-source image annotation tool employed to develop label files [13]. It is popular among many people because it supports several label files formats and it is user-friendly. In its application, by merely drawing around the target object using a rectangular box, the label file will be then automatically produced and saved in the desired pathway by LabelIMG. Prior to labeling we should prepare a text file that gives the calculations of the sequence numbers per category. It is numbered 0 onwards, e.g. 0 is H-1 and 1 is H-2 and so on. The label file is stored in the file with the extension of YOLO format, as illustrated, classes_index represents the type of object, x_center is the x-coordinate of the center point of the bounding box, y_center is the y-coordinate of the center point of the bounding box, w0 is the width of the bounding box, and h0 is the height of the bounding box.

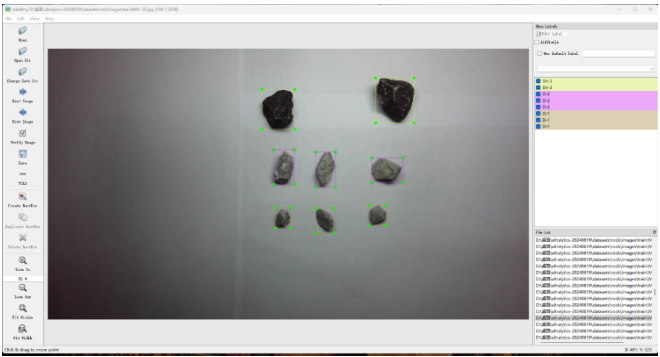

Fig. 2. LabelImg Tool Annotation Interface

Once the image annotation process has been done, a label file would be generated. The label file will be associated with each image. When the dataset is made, one will need to create a folder called images, in which the images will be put, and a folder called labels, in which the label files will be put. During storage, it must be made sure stringently that images and labels are ordered in correct order; failure to which, object detection will be out of control. In this case, you can make a sequential rearrangement and check with the help of the functions os.listdir().

## 3. Overview of the proposed robot prototype
### 3.1. Aggregate size measurement

Once the identification and positioning of the aggregate has been done, then another thing which must be done is to measure the aggregate particle size. In this article, the overall measurement of the particle size is approximately done. The lengths, width and diagonal lengths of the rectangle that is confined to the aggregate target detection are calculated according to the perimeter based on which the approximate size of the particles that form the aggregate is finally determined.

### 3.1.1 Particle size measurement algorithm implementation

When measuring aggregate size, we use an advanced measurement method based on a minimum enclosing rectangle [14]. The core principle is clear: once the aggregate is accurately identified and located, the system automatically constructs a minimum enclosing rectangle that fits the aggregate. This rectangle has a unique advantage. Not only can it completely and tightly wrap the aggregate, but its area is also relatively small among all rectangles that can wrap the aggregate, so it can fit the actual contour of the aggregate to the greatest extent.

First, with the help of professional and precise specific image recognition algorithms, the system can accurately capture the coordinate information points of the upper left corner and lower right corner of the minimum circumscribed rectangle [15]. The specific position and extension range of the rectangle in the two-dimensional plane are accurately determined, providing support for subsequent measurements.

Then the length and width parameters of the rectangle are calculated based on the obtained coordinate difference. In detail, the absolute value of the difference between the horizontal coordinates of the two coordinate points corresponds exactly to the length of the rectangle; and the absolute value of the difference between the vertical coordinates is exactly equal to the width of the rectangle. Once the values of the length and width are successfully obtained, use the Pythagorean theorem to add the square of the length and the square of the width and take the square root to get the length of the diagonal [16].

Given that the aggregate is always firmly contained within this carefully constructed rectangular frame, the geometric parameters of the rectangle itself, such as length, width and diagonal length, can naturally reflect the actual size characteristics of the aggregate. By comprehensively analyzing and rationally applying these key data, we can eventually successfully obtain the approximate particle size range of the aggregate, and thus achieve the task of roughly measuring the aggregate particle size [17, 18].

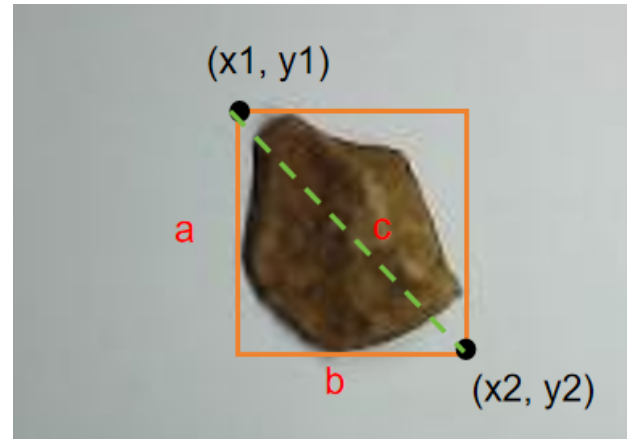

Fig. 3. Schematic diagram of particle size measurement

As shown in Fig. 3, the length, width and diagonal length of the aggregate are a, b and c respectively, and the calculation method is as follows:

$$a = |y1 - y2|$$
$$b = |x1 - x2| \qquad (1)$$
$$c = \sqrt{a^2 + b^2}$$

After calculating the length, width and diagonal dimensions, the particle size of the aggregate can be roughly determined and the aggregate can be roughly measured.

### 3.2. Software System

The system used in this study is Ubuntu 20.04 [19, 20]. Ubuntu has distinct and advantageous features. The first is ease of use. Its graphical user interface is simple and intuitive, the menu layout is reasonable, and the icons are clear and easy to understand. New users can quickly become familiar with the operation, such as easily finding the launch entry for commonly used software. Software installation is even more convenient. The Ubuntu Software Center is like a convenient supermarket. You can install various software such as office and entertainment, with one click by entering the software name. The stability performance is excellent. The kernel has been repeatedly optimized and can run smoothly for a long time even under high load. The server can be deployed for several months without restarting. Moreover, it has good hardware compatibility and is compatible with mainstream hardware brands and models, lowering the threshold for use. The software is extremely rich, and the huge software warehouse covers a wide range of fields from basic office to cutting-edge scientific research, from daily entertainment to professional development, providing sufficient software resource support for all types of users.

The robot arm is driven and controlled using the ROS system, the Robot Operating System, which is a key force in the field of robot development. It was born in the open source community and brings together the wisdom of developers around the world. ROS is highly flexible and adopts a distributed architecture, allowing different functional components to run in the form of nodes. Each node can work independently and collaborate through a message passing mechanism, which makes the development of complex robot systems clear and organized.

For example, in an autonomous vehicle project, modules such as perception, decision-making, and control act as independent nodes and interact efficiently. Its rich feature package is a treasure trove, covering many areas such as robot motion control, visual processing, map construction, etc. Developers do not need to start from scratch, which greatly saves R&D time. At the same time, supporting visualization tools such as RViz can intuitively present the robot's operating status, making it easier to debug and optimize. The programming language used is Python 3.8 [21].

### 3.3. Hardware system

The hardware equipment used in this article mainly includes a six-degree-of-freedom robotic arm, the robotic arm product is Hiwonder's JetArm robotic arm, an Orbbec Gemini binocular camera, a computer, a rigid robotic claw, a ROS motherboard, and an electronic control system.

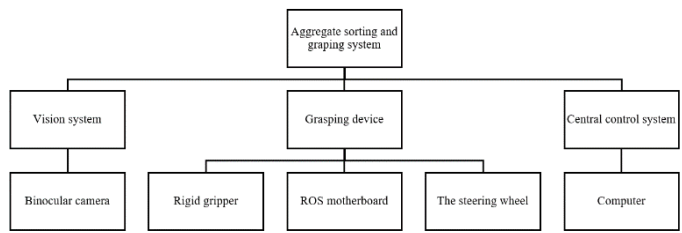

Fig. 4. System hardware components

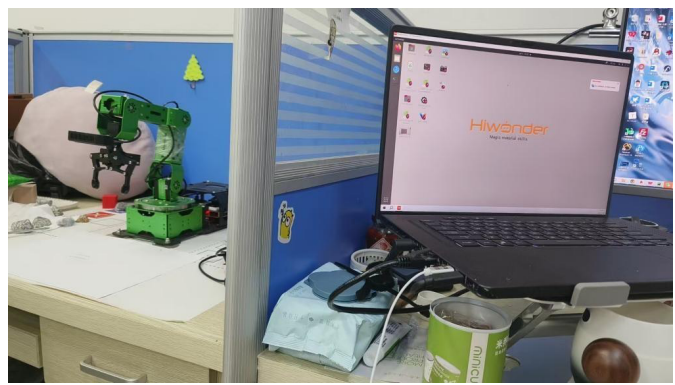

Fig. 5. Hardware layout

(1) Binocular camera

The binocular camera used is the Gemini camera produced by Orbbec [22]. The Gemini embedded 3D camera module can obtain depth images and color images of objects. It has the advantages of small size and high performance. It is suitable for 3D object scanning terminal devices with a recognition distance of 25-250cm and supports USB3.0 interface. Using active binocular structured light design, Gemini can provide high-resolution depth maps and infrared image information with strong light resistance. From Fig. 4, we can see that the left and right cameras of the binocular camera are infrared cameras, and there is also a color camera, which can obtain RGB color images while obtaining infrared depth maps. The infrared projector is used to project structured light. After the structured light interacts with the surface of the object, it is captured by the infrared camera. The depth information of the object is obtained by calculating the difference in the structured light images obtained by the left and right cameras. The binocular camera can not only provide three-dimensional coordinate information of objects, but also realize object recognition and tracking in combination with color images, providing key data support for the robot arm's precise grasping and placement tasks.

The parameters of the binocular camera are shown in the following table:

Table 1

| Binocular Camera | | |
|---|---|---|
| Infrared camera | | Color camera |
| Working scope | 0.25~2.5m | UVC support |
| Accuracy | 1m：±5mm | |
| Resolution@Framerate | 1280×800@30fps | 1920×1080@30fps |
| | 640×400@60fps | 640×480@60fps |
| Deep processing chip | MX6000 | |
| Data Interface | | USB3.0 |
| Supported operating systems | | Windows/Linux/Android |

(2) Robotic arm and control components

The robotic arm in this article is the JetArm robotic arm produced by Hiwonder. JetArm is a three-link intelligent robotic arm with deep learning and computer vision capabilities [23, 24]. It is equipped with NVIDA JetsonNano, 3D depth camera, 6DOF robotic arm and other high-performance intelligent hardware configurations. The built-in inverse kinematics algorithm realizes multi-scene, high-precision AI recognition and precise gripping and placement functions. It can perform sorting, stacking, classification and other gameplay. If equipped with optional hardware such as microphone and sound card, it can also realize voice control of robotic arm color recognition and tracking gameplay.

The robot electronic control uses the STM32 controller as the underlying motion controller to complete the communication between the Jeson mainboard and the bus servo. The controller uses the STM32F407VET6 [25] as the main control, which uses the ARM Cortex-M4 core [26], 168M main frequency, 512K on-chip FLASH capacity, 192K on-chip SRAM capacity, integrated FPU and DSP instruction control board with rich on-board resources and expansion interfaces, which is very suitable for ROS robot development, and can form a ROS robot with the Jetson series ROS main control.

The power supply is a 12V5A adapter [27]. After the battery 12V is input into the STM32 controller, it directly powers the bus servo. After DCDC, it is converted into a 5V voltage to power the JetsonNano main control [28]. The maximum current is 5A, which can well meet the power supply requirements of Jetson Nano.

The robot uses Jetson Nano as the ROS master control, which consists of a Jetson Nano mainboard + a Jetson mini expansion board. The mainboard is a small but powerful computer that can run mainstream deep learning frameworks and can meet the computing power required for most artificial intelligence projects. The Ubuntu 18.04 system is installed on the mainboard, and the robot operating system ROSMelodic environment is built.

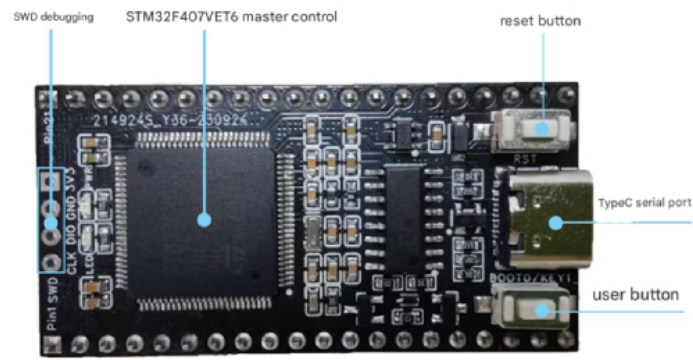

Fig. 7. Control panel

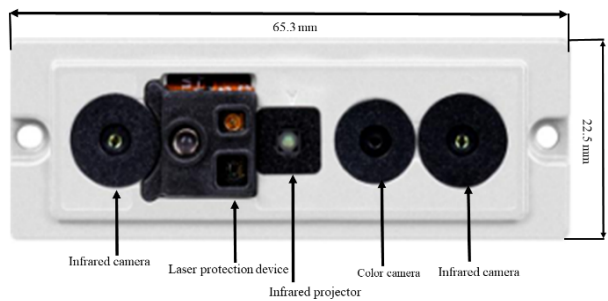

Fig. 6. Gemini stereo camera

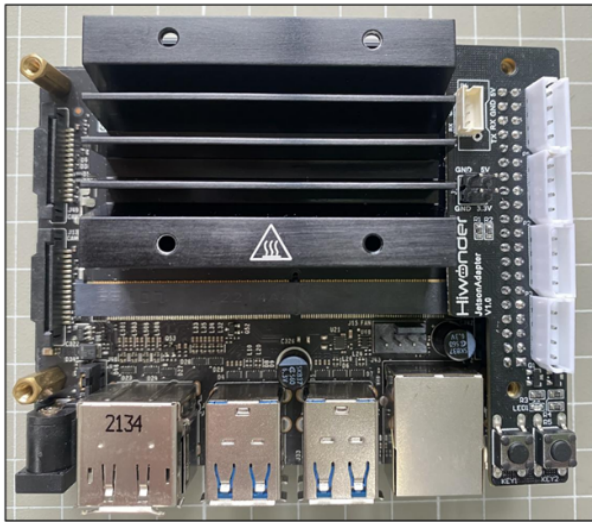

Fig. 8. Main controller

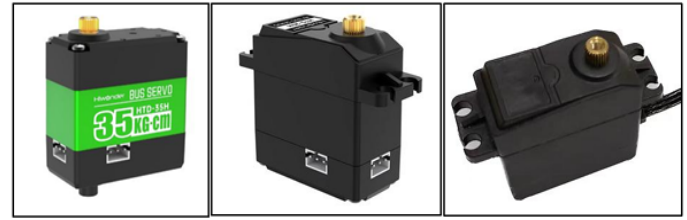

Fig. 9. Servo

JetArm is a 6-DOF robotic arm, which is composed of 6 intelligent bus servos and metal sheet metal parts. The bus servos use serial communication to connect multiple servos through a bus. The bus servos can connect multiple servos in series through an I/O port, thus achieving the characteristics of "high precision and slightly more expensive than digital servos".

### 3.4. Real machine construction and experiment

After completing the selection of hardware and software, the next step is to build the entire platform system. All the previous work needs to be connected in series as a whole so that the system can run smoothly. Then it is necessary to conduct a physical grasping experiment to verify the feasibility of the system. This section will explain the system construction and physical experiments in detail.

#### 3.4.1. Real machine platform construction

The construction of the aggregate recognition and classification system mainly consists of three parts: target detection module, positioning module and grasping module. First, a data set for training is created in the target detection module, and then the YOLOv8 algorithm is improved and verified [29]. The model accuracy is improved by adding an attention mechanism, improving the C2f module [29], and improving the detection head. In the positioning module, the camera's internal and external parameters and hand-eye calibration are first performed to obtain an accurate hand-eye transformation relationship. The left and right images are then captured using a binocular camera. The stereo matching algorithm is then improved, and Census transformation and gradient cost are added to the cost calculation to achieve better small target detection effects, obtain the depth map of the aggregate, and realize three-dimensional positioning of the aggregate. In the grasping module, the DH modeling and kinematic analysis of the robot arm were first completed to make the control of the robot arm more precise. The hardware equipment was selected and built to obtain the aggregate classification and identification system.

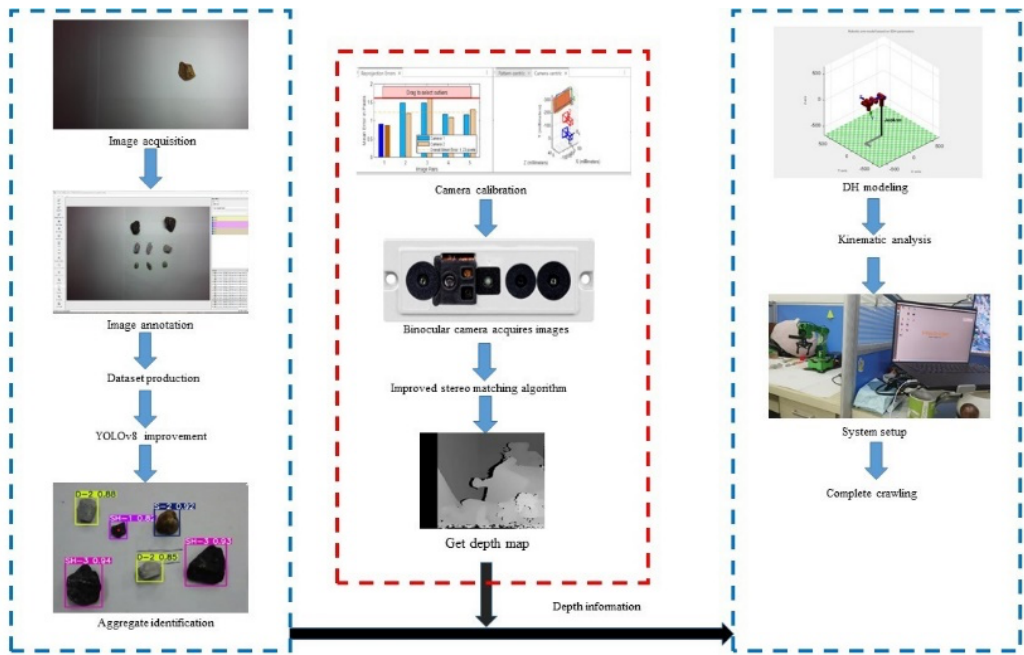

Fig. 10. Platform construction

**3.5. Crawl Experiment and Analysis**

This section will introduce the actual experiment. The experimental process is shown in the figure below:

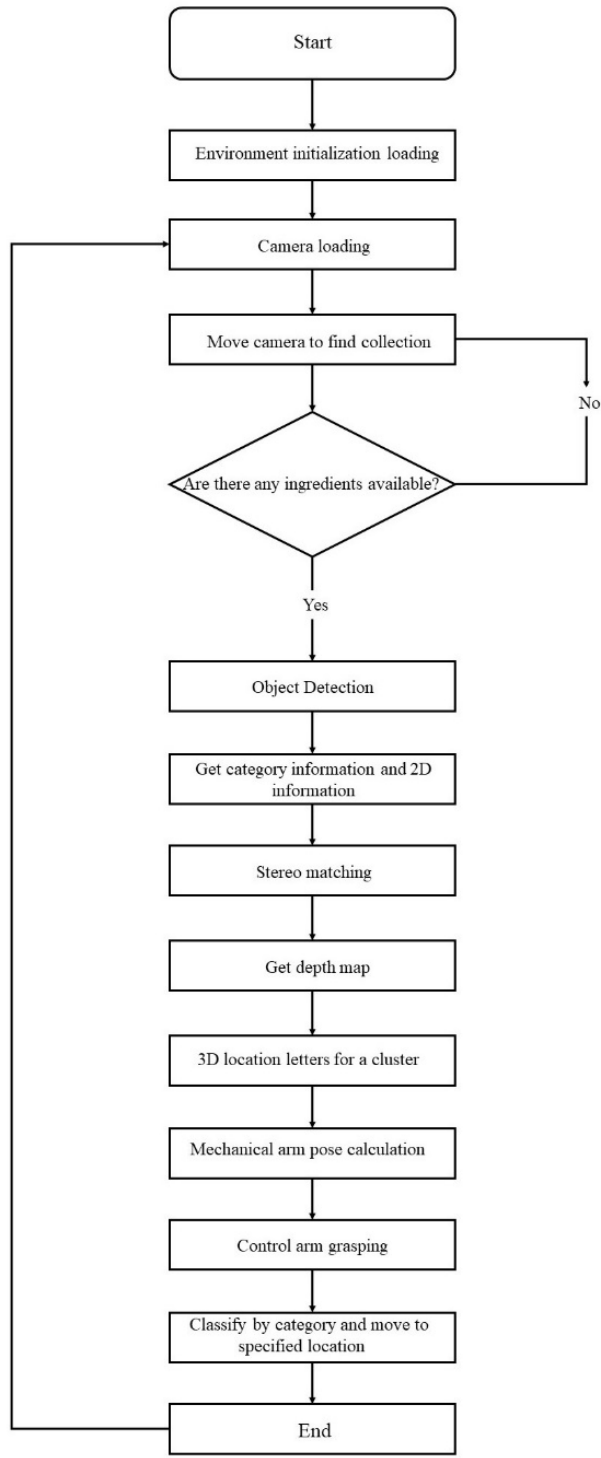

Fig. 11. Experimental flow chart

First, load the environment. Since it is remotely controlled through a computer, you need to load the virtual environment on the computer first, then set various parameters and complete the initialization. Then load the binocular camera, check the imaging effect of the binocular camera, and then move the binocular to search for aggregates. After finding the target object, perform the target detection stage, otherwise continue searching until the target object appears in the field of vision. Target detection first loads the trained weight file, calls the weight file for target detection, realizes the lithology classification of aggregates [30], and obtains the two-dimensional position information of aggregates. Then load the left and right infrared cameras of the binocular camera, acquire images, perform stereo matching to obtain the depth map, obtain the depth information of the aggregate, and combine the previous two-dimensional position information to obtain the three-dimensional position of the aggregate. The aggregate particle size measurement first finds the minimum circumscribed rectangle of the aggregate, calculates the length and width information of the aggregate, and obtains the actual particle size based on the three-dimensional information. After acquiring the three-dimensional information, the position and posture of each axis of the robot arm are calculated according to the end position, and the robot arm is driven to reach the designated position according to the position and posture to grab the aggregate. Then, according to the lithology type of the grabbed aggregate [31], the aggregate is placed at the pre-set position. After the above steps, the task of grabbing an aggregate is completed, and then the above steps are repeated to complete the identification and grabbing tasks of the entire experiment.

First, we need to complete the target detection of aggregates. We have completed the data set preparation, YOLOv8 algorithm improvement and algorithm performance verification [29]. We put various aggregates into the recognition range and drive the binocular camera to recognize the aggregates. The recognition effect is shown in the figure below. It can be seen that the recognition result of the aggregate has a high confidence level, and the recognition accuracy of different categories is close to 100%.

After completing the key step of aggregate identification and detection, the focus of subsequent operations is on how to accurately drive the robotic arm to grab the aggregate. The core of this process lies in coordinate transformation and information transmission. First, coordinate transformation is achieved based on the results of hand-eye calibration. Hand-eye calibration can determine the relative position relationship between the binocular camera and the robotic arm, based on which the coordinates acquired by the binocular camera are converted into the coordinates of the robotic arm. This conversion process is crucial as it provides the robot with precise end coordinate information, allowing the robot to accurately know the location of the aggregate. Next, the obtained robot arm end coordinate information is transmitted to the ROS mainboard [32]. As the "brain" of the entire system, the ROS mainboard undertakes complex computing tasks. After receiving the data from the coordinate transformation, it will accurately calculate the angle that each servo needs to rotate according to the structure and kinematic model of the robotic arm. Finally, the calculated angle data will be sent to the corresponding servos. After receiving the command, the servo immediately rotates. Through the coordinated rotation of each servo, the end of the robotic arm can accurately reach the location of the aggregate, thereby completing the grabbing action of the aggregate. The entire process is closely linked, from coordinate conversion to information transmission, and then to servo control. Each link relies on precise calculation and stable communication. Only in this way can the efficiency and accuracy of the robotic arm in grabbing aggregates be ensured, realizing this key operation in the automated production process.

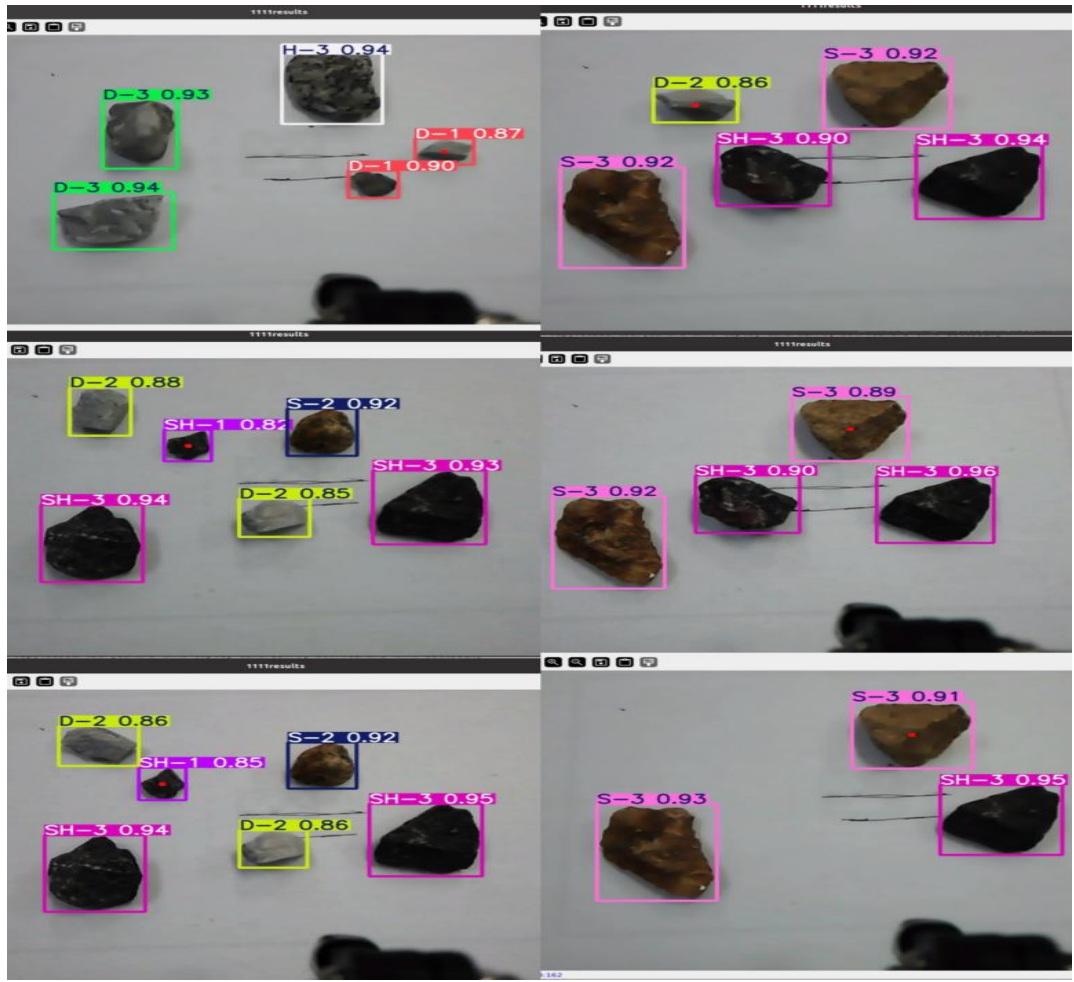

Fig. 12. Target detection effect diagram

The crawling process is shown in the following figure:

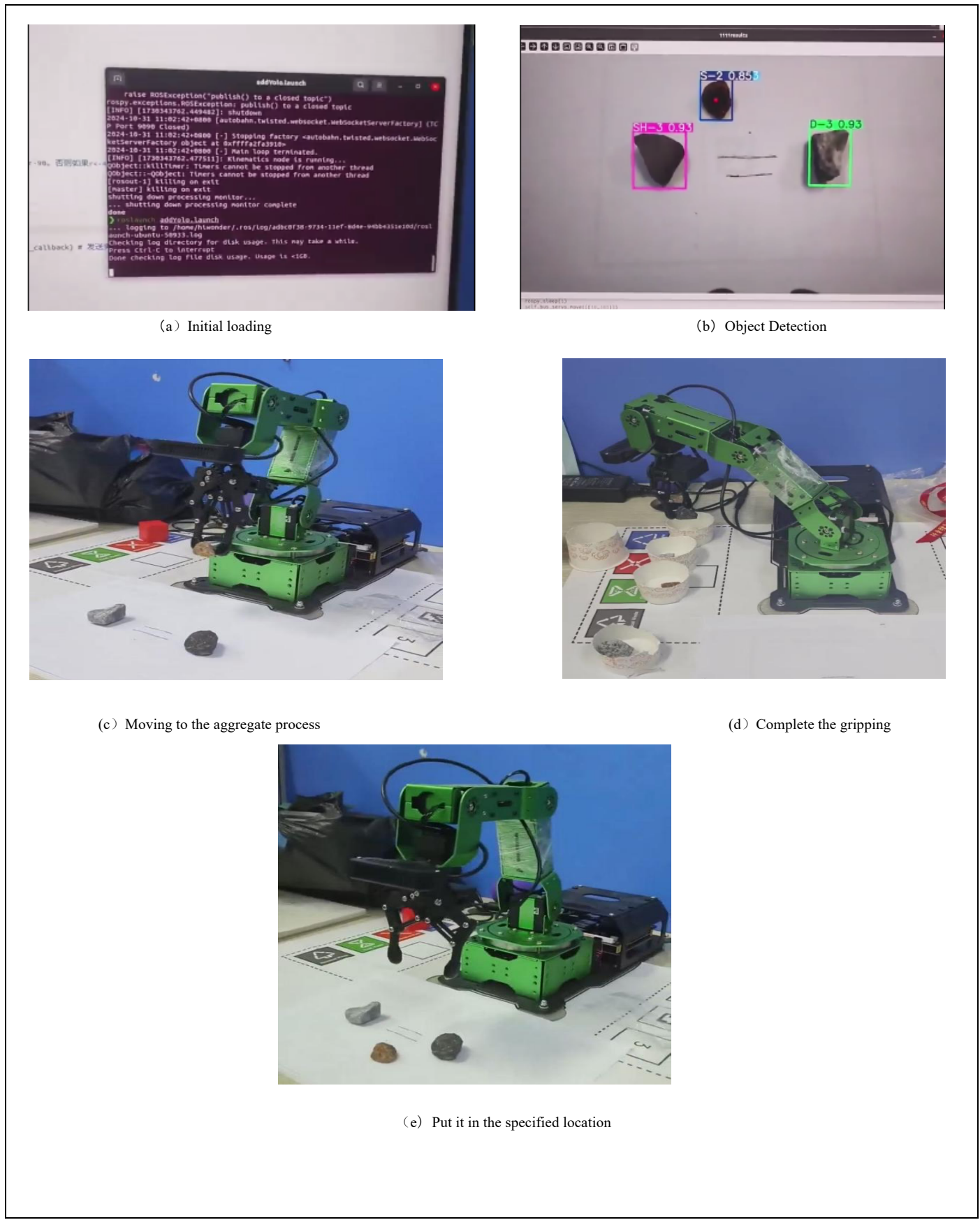

(a) Initial loading

(b) Object Detection

(c) Moving to the aggregate process

(d) Complete the gripping

(e) Put it in the specified location

Fig. 13. Capture process

Md. Taherul et al.

There are four types of rocks in this paper, namely limestone, marble, sandstone and granite [33]. Therefore, four groups of grasping experiments are designed to verify the accuracy of grasping and classification. Each group conducts 10 graspings. Each type of rock includes three different particle size grades. Mixed aggregates of various particle sizes are randomly selected during selection. The grasping success rate, that is, the robot arm successfully reaches the aggregate position and completes the grasping without falling, and the classification success rate, that is, the correct rate of each type being placed in the specified position, are recorded. The experimental results are shown in the following table:

| Category | Grab aggregate quantity/particle | Number of successfully captured particles | Success rate/% | Correctly classified quantity/piece | Accuracy/% |
|---|---|---|---|---|---|
| limeston（SH） | 10 | 10 | 100 | 10 | 100 |
| granite（H） | 10 | 10 | 100 | 9 | 90 |
| Sandstone (S) | 10 | 9 | 90 | 10 | 100 |
| marble（D） | 10 | 10 | 100 | 10 | 100 |

From the data results of the grasping experiment, it can be seen that the system has a good performance in terms of success rate, with a success rate close to 100%. However, when grasping 1 cm aggregate, it may fail to clamp. This is because the aggregate particle size is too small and the gripper does not have a good fulcrum when grasping, resulting in failure to grasp small-size aggregate. Another failure was caused by the system misidentifying granite as limestone, which occurs when the aggregate size is small and the surface texture is not obvious. The jaws need to be improved in the future. The inclination angle of the jaws can be changed to make the jaws better contact with the aggregate plane and thus have better force. At the same time, further improvements can be made in the target detection algorithm to improve the detection effect of small targets.

## 4. Key methodology
### 4.1. Robotic arm posture description

To achieve accurate grasping of the robot arm, the position and posture of the robot arm must be accurately analyzed. When the robot arm is grasping, the axes actually move relative to each other so that the end can accurately reach the location of the target object. When studying the position and posture of the robot arm, the world coordinate system usually coincides with the coordinate system of the robot arm base, and a point P in space can be represented by a vector:

$$P = \begin{bmatrix} p_x p_y p_z \end{bmatrix}^T \quad (4.1)$$

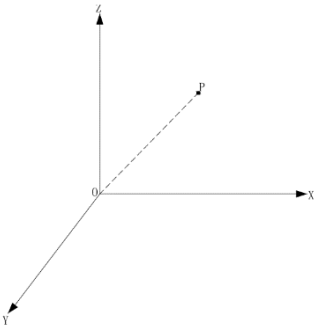

Fig. 14. Position representation

When studying the position and posture of the robot arm, a corresponding coordinate system is usually established on each axis. Each coordinate system can be described by transforming the coordinate system of the base. First, the $O_1$-$X_B Y_B Z_B$ coordinate system is obtained by rotation [34], as shown in the figure:

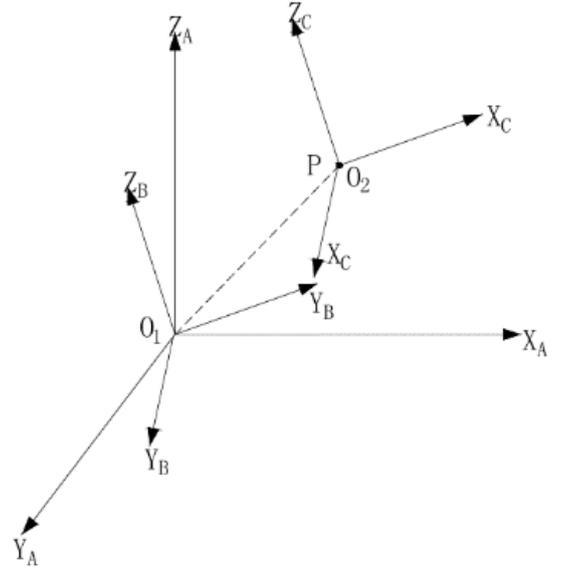

Fig. 15. Schematic diagram of pose transformation

$$R = \begin{bmatrix} {}^A X_B & {}^A Y_B & {}^A Z_B \end{bmatrix} = \begin{bmatrix} r_{11} & r_{12} & r_{13} \\ r_{21} & r_{21} & r_{23} \\ r_{31} & r_{32} & r_{33} \end{bmatrix} \quad (4.2)$$

When the origins of the two coordinate systems do not coincide, they can be transformed by rotation and translation, as shown in Fig. 15. After rotation and then translation, the transformation of the coordinate system can be expressed as:

$$\{O_2\} = \{R \quad P\} \quad (4.3)$$

### 4.2 D-H model and kinematic analysis

D-H modeling describes the transformation from one coordinate system to another by defining a standard set of parameters. With these transformations, kinematic analysis can be simplified to aligning the spatial relationships between the robot joints and the end effector [35]. D-H usually has four parameters $a_i$, namely, connecting rod length $\alpha_i$, connecting rod rotation angle $\theta_i$, joint angle $\mathbf{d}_i$, and connecting rod offset. The definitions of each parameter are as follows:

$a_i$ — represents the displacement of the adjacent coordinate system along the x-axis of the previous coordinate system;

$\alpha_i$ —Indicates the angle of rotation of the adjacent coordinate system around the x-axis of the previous coordinate system;

$\theta_i$ —Indicates the angle of rotation of the adjacent coordinate system around the z-axis of the previous coordinate system;

$\mathbf{d}_i$ —Represents the displacement of the adjacent coordinate system along the z-axis of the previous coordinate system.

Based on the D-H parameters, the transformation matrix between the two coordinate systems is:

$$T_i^{i-1} = \begin{bmatrix} \cos(\theta_i) & -\sin(\theta_i) & 0 & a_{i-1} \\ \sin(\theta_i)\cos(\alpha_{i-1}) & \cos(\theta_i)\cos(\alpha_{i-1}) & -\sin(\alpha_{i-1}) & -\sin(\alpha_{i-1})d_i \\ \sin(\theta_i)\sin(\alpha_{i-1}) & \cos(\theta_i)\sin(\alpha_{i-1}) & \cos(\alpha_{i-1}) & \cos(\alpha_{i-1})d_i \\ 0 & 0 & 0 & 1 \end{bmatrix}$$

(4.4)

Through the above transformation matrix, the position transformation relationship between each link can be determined, which is convenient for the subsequent kinematic analysis of the robot arm. The D-H parameters of the robot arm in this article are shown in the following table:

Table 2

| i | $\theta_i/°$ | $d_i/m$ | $a_i/m$ | $\alpha_i/°$ |
|---|---|---|---|---|
| 1 | $\theta_1$ | 0 | 0 | 0 |
| 2 | $\theta_2$ | 0 | 0 | -90 |
| 3 | $\theta_3$ | 0 | 0.1294 | 0 |
| 4 | $\theta_4$ | 0 | 0.1294 | 0 |
| 5 | $\theta_5$ | 0 | 0 | -90 |

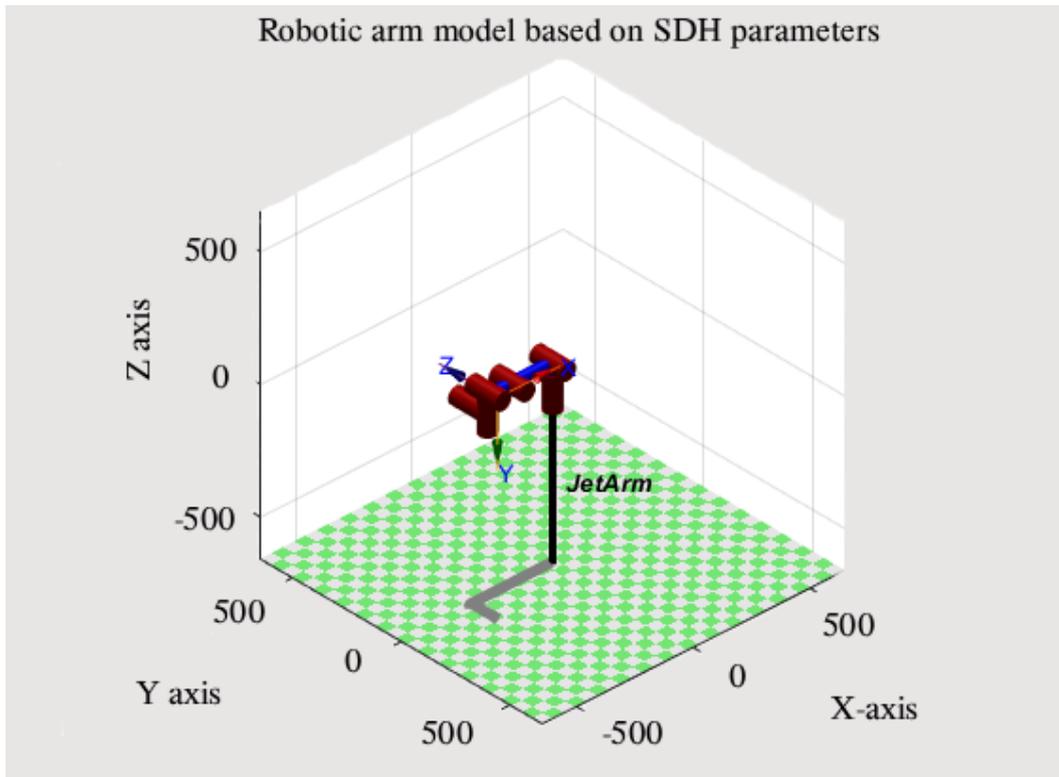

Fig. 16. Robotic arm D-H modeling

The purpose of solving the forward kinematics of the robotic arm is to solve the spatial position and posture of the end effector by knowing the variables of each joint. By constructing the coordinate transformation matrix of each joint and multiplying the transformation matrix of each joint, the position and posture of the end effector relative to the base are finally obtained.

Coordinate transformation from base to end effector:

$$^0_5T = {}^0_1T\,{}^1_2T\,{}^2_3T\,{}^3_4T\,{}^4_5T \tag{4.5}$$

Substituting the D-H parameters into equation (5.5) yields:

$$^0_1T = \begin{bmatrix} \cos(\theta_1) & -\sin(\theta_1) & 0 & 0 \\ \sin(\theta_1) & \cos(\theta_1) & 0 & 0 \\ 0 & 0 & 1 & 0 \\ 0 & 0 & 0 & 1 \end{bmatrix}, \quad {}^1_2T = \begin{bmatrix} \cos(\theta_2) & 0 & \sin(\theta_2) & 0 \\ 0 & 1 & 0 & 0 \\ -\sin(\theta_2) & 0 & \cos(\theta_2) & 0 \\ 0 & 0 & 0 & 1 \end{bmatrix}$$

$$^2_3T = \begin{bmatrix} \cos(\theta_3) & -\sin(\theta_3) & 0 & 0 \\ \sin(\theta_3) & \cos(\theta_3) & 0 & 0 \\ 0 & 0 & 1 & 0 \\ 0 & 0 & 0 & 1 \end{bmatrix}, \quad {}^3_4T = \begin{bmatrix} \cos(\theta_4) & -\sin(\theta_4) & 0 & 0 \\ \sin(\theta_4) & \cos(\theta_4) & 0 & 0 \\ 0 & 0 & 1 & 0 \\ 0 & 0 & 0 & 1 \end{bmatrix} \tag{4.6}$$

$$^4_5T = \begin{bmatrix} \cos(\theta_5) & 0 & \sin(\theta_5) & 0 \\ 0 & 1 & 0 & 0 \\ -\sin(\theta_5) & 0 & \cos(\theta_5) & 0 \\ 0 & 0 & 0 & 1 \end{bmatrix}$$

$$^0_5T = {}^0_1T\,{}^1_2T\,{}^2_3T\,{}^3_4T\,{}^4_5T = \begin{bmatrix} n_x & o_x & a_x & p_x \\ n_y & o_y & a_y & p_y \\ n_z & o_z & a_z & p_z \\ 0 & 0 & 0 & 1 \end{bmatrix} \tag{4.7}$$

$n_x = c_1 c_5 + s_1 s_5$
$n_y = s_1 c_5 - c_1 s_5$
$n_z = 0$
$o_x = -c_1 s_5 + s_1 c_5$
$o_y = -s_1 s_5 - c_1 c_5$
$o_z = 0$
$a_x = 0$
$a_y = 1$
$a_z = -1$
$p_x = c_1 (c_{23} a_3 + c_2 a_2 + a_4)$
$p_y = s_1 (c_{23} a_3 + c_2 a_2 + a_4)$
$p_z = -d_5 - s_{23} a_3 - s_2 a_2$

Among them, $s_i = \sin \theta_i$, $c_i = \cos \theta_i$, $\begin{bmatrix} p_x & p_y & p_z \end{bmatrix}^T$, represents the position of the end of the robot arm in the base coordinate system. After the forward kinematics solution, its correctness needs to be verified. According to the D-H parameters, the Robotics Toolbox toolbox is used in MATLAB for verification. For any given set of joint angles $\theta = \begin{bmatrix} 90 & 0 & 0 & 0 & 0 \end{bmatrix}$, the kinematic solution result is:

$$^0_5T = \begin{bmatrix} 0 & 0 & 1 & 0.2588 \\ 1 & 0 & 0 & 0 \\ 0 & 1 & 0 & 0 \\ 0 & 0 & 0 & 1 \end{bmatrix} \tag{4.8}$$

Substituting the joint angle into equation (5.6), the result is the same as the above result, which verifies the accuracy of the forward kinematics solution.

Inverse kinematics analysis is the process of solving the angle or position that each joint should have given the desired position and posture of the end effector. In the control of the robotic arm, the goal is usually to achieve the specified task by controlling the position and posture of the end effector. Therefore, the purpose of inverse kinematics is to calculate the specific angle or displacement of each joint of the robotic arm based on the given target end position and desired posture.

According to the previous formula, the joint angles can be obtained as:

$$\theta_1 = \arctan(P_y / P_x) = \{c_1(c_{23}a_3 + c_2 a_2 + a_4) / s_1(c_{23}a_3 + c_2 a_2 + a_4)\}$$

$$\theta_2 = \arcsin\left(\begin{array}{c} -(P_z + d_5) / \sqrt{((c_3 a_2)*(c_3 a_2) + (s_3 a_3)*(s_3 a_3))} \\ -\arctan(s_3 a_3 / c_3 a_3 + a_2) \end{array}\right)$$

$$\theta_3 = \pm \arccos\left(\begin{array}{c} (P_z^2 + P_y^2 + P_x^2 - 2a_4(c_1 P_x + s_1 P_y)) \\ +2P_x d_5 + a_4^2 + d_5^2 - a_2^2 - a_3^2 \end{array}\right) / (2a_2 a_3) \tag{4.9}$$

$$\theta_4 = -(\theta_2 + \theta_3)$$

$$\theta_5 = \theta_1 - \arctan(n_y / n_x)$$

After the calculation is completed, MATLAB is used to verify the solution and the actuator end pose matrix is randomly given:

$$^0_5T = \begin{bmatrix} 0 & 0 & 1 & 0.2588 \\ 1 & 0 & 0 & 0 \\ 0 & 1 & 0 & 0 \\ 0 & 0 & 0 & 1 \end{bmatrix} \tag{4.10}$$

After solving, a set of optimal results can be obtained: $\theta = \begin{bmatrix} 90 & 0 & 0 & 0 & 0 \end{bmatrix}$,

Compared with the solution results, it can be seen that the calculated results are basically consistent, which verifies the correctness of the inverse kinematics solution.

## 5. Experiments

### 5.1. Binocular Camera Calibration Experiment

This section will calibrate the binocular camera. The binocular camera used is the Gemini plus camera produced by Orbbec. The calibration method is the Zhang Zhengyou calibration method [36, 37]. The camera is calibrated using MATLAB software [38]. The calibration tool used in the calibration process is a 9×6 black and white chessboard, as shown in Fig. 17. The side length of the chessboard is 27 mm.

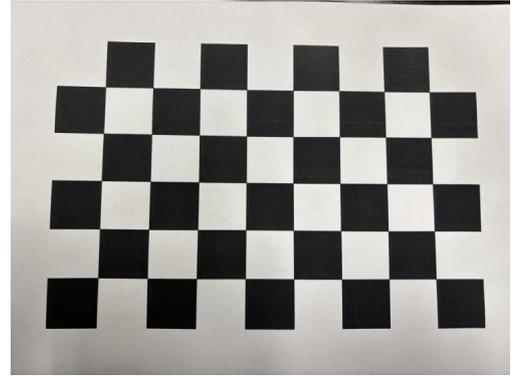

Fig. 17. Checkerboard calibration

Image acquisition: First, use a binocular camera to take pictures of the calibration plate from different angles. Fix the camera, move the calibration plate, and collect 20 pictures of the calibration plate, as shown in Fig. 18. As can be seen from the figure, since the left and right cameras of the camera are both infrared imaging, the pictures taken are black and white, and the image quality is not high. It is easily disturbed by noise. The corners of the image are not very clear and it is easy to cause recognition errors. Therefore, the taken pictures are enhanced and processed by PS. The processed pictures are shown in Fig. 19. Through processing, the corners of the chessboard are more obvious and more conducive to corner detection.

Corner point extraction: After the acquisition is completed, the next step is to extract the corner points. The corner points are located at the intersection of the black and white grids [39]. The origin of the world coordinate system is located at the first corner point in the lower right corner of the image, marked with an orange box. The x-axis is positive horizontally to the left and the y-axis is positive vertically upward. The corner points are marked with green circles, as shown in Fig. 20.

*Md. Taherul et al.*

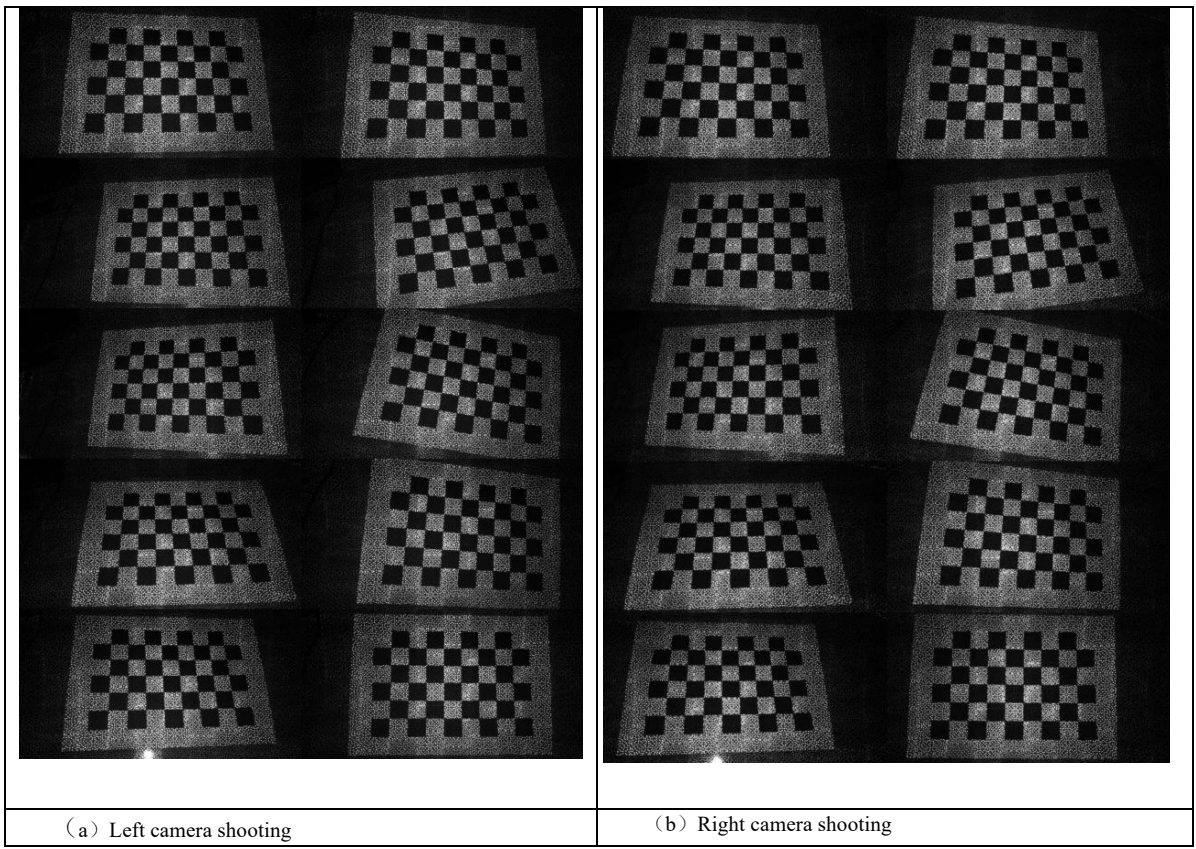

| （a）Left camera shooting | （b）Right camera shooting |

Fig. 18. Left and right camera acquisition images

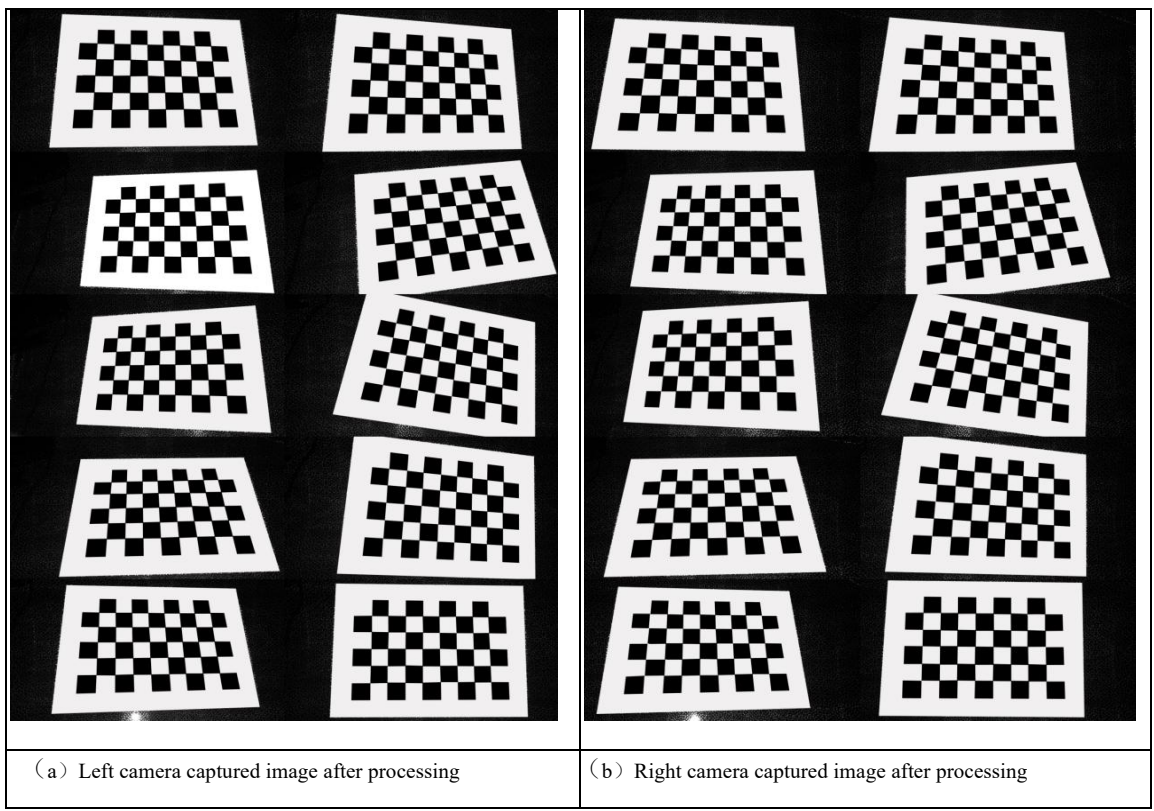

| （a）Left camera captured image after processing | （b）Right camera captured image after processing |

Fig. 19. The collected image after processing

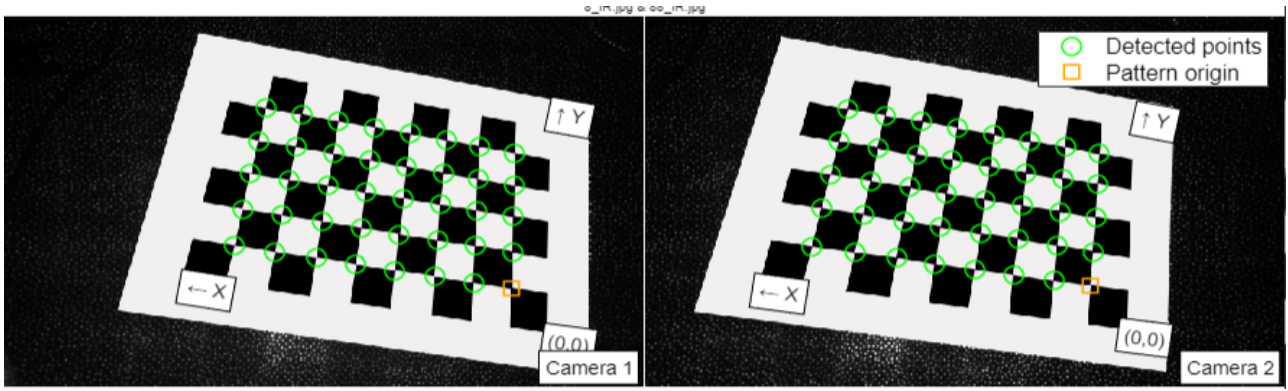

Fig. 20. Corner point extraction

Camera calibration: After corner point extraction, the Stereo Camera Calibrator tool in MATLAB [40] can be used to automatically calculate parameters, including the internal and external parameters of the camera and distortion parameters [41]. During calibration, remove images with large errors until the final error meets the requirements. After the final adjustment, the pixel reprojection error is 1.23, which is less than 1.5, indicating that the accuracy of the calibration method meets the experimental requirements. The final calibration results are shown in the following table:

| Parameter | Left camera | Right camera |
|---|---|---|
| Internal parameter matrix | $\begin{bmatrix} 22.9741 & 0 & 405.0067 \\ 0 & 30.9439 & 430.6781 \\ 0 & 0 & 1 \end{bmatrix}$ | $p_c = [X_c \; Y_c \; Z_c \; 1]^T \; p_E = [X_E \; Y_E \; Z_E \; 1]^T \; p_B = [X_B \; Y_B \; Z_B \; 1]^T$ $\begin{bmatrix} 25.3241 & 0 & 443.4829 \\ 0 & 28.7749 & 437.5979 \\ 0 & 0 & 1 \end{bmatrix}$ |
| Distortion coefficient | $K_1 = -0.0000716 \; k_2 = 0.00000003$ $P_1 = 0.0016 \; p_2 = -0.000247$ | $K_1 = 0.00023 \; k_2 = -0.00000005$ $P_1 = 0.0012 \; p_2 = -0.000252$ |
| Camera rotation matrix | $\begin{bmatrix} 0.9991 & -0.0041 & -0.0019 \\ 0.0041 & 0.9991 & 0.0019 \\ 0.0014 & -0.0120 & 0.9999 \end{bmatrix}$ | |
| Camera translation matrix | $[-4.9553 \; 121.2709 \; 5.1484]$ | |

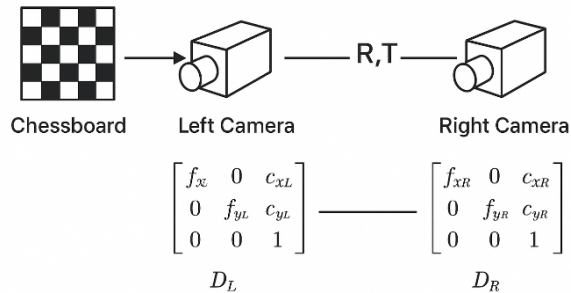

Fig. 21. Schematic diagram of camera calibration results

### 5.2. Hand-eye calibration

The hand-eye calibration of the system is also the calibration between the robot arm and the camera [42, 43]. The camera has been calibrated in the previous article to realize the extraction and conversion of the three-dimensional coordinates of the object. When performing the grasping operation, the robot arm needs to move to the specified position for grasping. Therefore, the coordinates of the robot arm also need to be included in the entire coordinate system to achieve accurate grasping of the robot arm. There are two main forms of positional relationship between the robotic arm and the camera, namely 'eyes on the hand' and 'eyes outside the hand', as shown in Fig. 17. "Eyes on the hands" means that the camera is fixed on the end of the robotic arm.

The camera will move with the end of the robotic arm, and the recognition range of the camera will change with the end of the robotic arm. In this way, the camera's field of view can be changed according to the needs of the robotic arm. During calibration, the camera coordinate system and the robotic arm's coordinate system need to be calibrated. "Eyes outside the hand" means installing the camera at a fixed location outside the robotic arm [44]. The camera's field of view will always be fixed within a specific range. This method does not require frequent changes in the field of view. During calibration, only the camera needs to be calibrated, and there is no need to calibrate the end section of the robotic arm. The setup chosen for this article is the 'eyes on hands' approach.

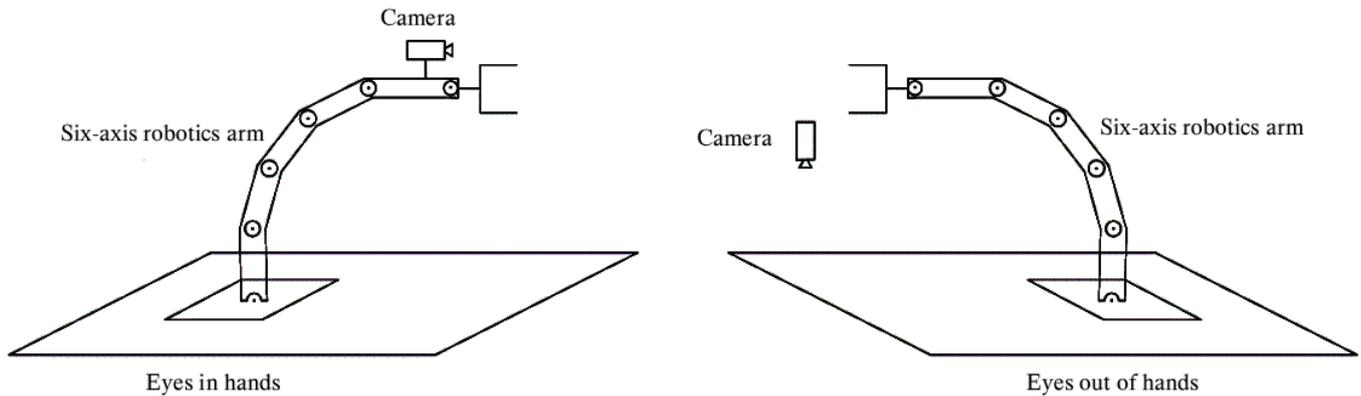

Fig. 22. Relative positions of the robot arm and camera

Define the camera coordinate system as C, the origin is the optical center of the camera, the coordinate axis direction is consistent with the camera line of sight, the robot end effector coordinate system is E, the origin is the center of the operating tool, the coordinate axis direction is consistent with the tool working direction, the robot base coordinate system is B, and its secondary coordinates are expressed as,

$$p_c = [X_c \ Y_c \ Z_c \ 1]^T \ p_E = [X_E \ Y_E \ Z_E \ 1]^T \ p_B = [X_B \ Y_B \ Z_B \ 1]^T$$

Transformation $T_{CE}$ of the camera relative to the end effector of the robot arm:

$$T_{CE} = \begin{bmatrix} R_{CE} & t_{CE} \\ 0 & 1 \end{bmatrix} \quad (5.1)$$

Where:
$R_{CE}$ —represents the rotation matrix;
$t_{CE}$ —represents the translation vector.
Transformation of the final actuator relative to the base $T_{EB}$:

$$T_{EB} = \begin{bmatrix} R_{EB} & t_{EB} \\ 0 & 1 \end{bmatrix} \quad (5.2)$$

Where:
$R_{EB}$ —rotation matrix of the end effector relative to the base;
$t_{EB}$ —Translation vector of the end effector relative to the base.
From the camera coordinate system to the end effector coordinate system:

$$p_E = T_{CE}^{-1} \cdot p_c \quad (5.3)$$

From the final actuator to the base coordinate system:

$$p_B = T_{EB}^{-1} \cdot p_E \quad (5.4)$$

From the camera coordinate system to the base coordinate system:

$$p_B = T_{CE}^{-1} \cdot \left( T_{CE}^{-1} \cdot p_c \right) = \left( T_{EB} \cdot T_{CE} \right)^{-1} \cdot p_c \quad (5.5)$$

When performing hand-eye calibration, the rotation matrix and translation vector are solved through the posture of the robot arm and the coordinates of the feature points recognized by the camera, and a series of postures of the robot end effector (including rotation $R_E$ and displacement $t_E$) and the position $T_c$ of the feature points seen by the camera in the camera coordinate system are collected. Then, by constructing the linear equation $T_{CE} = [R_{CE} | t_{CE}]$, R and t are obtained using the least squares method.

By continuously adjusting the position of the robot arm to obtain multiple sets of data, the hand-eye calibration matrix is solved

$$\begin{bmatrix} -9.7357e-1 & -1.0229e-16 & 2.2835e-1 & 1.0798e-1 \\ -1.7453e-16 & 1 & -2.9614e-16 & 3.1884e-1 \\ -2.2835e-1 & -3.2818e-16 & -937357e-1 & 1.9752e-1 \\ 0 & 0 & 0 & 1 \end{bmatrix} \quad (5.6)$$

Conclusion:
The paper proposes an automated aggregate sorter the aid of computer vision of a robotic arm system that goes beyond the constraints of the manual and mechanical signature that include inefficient manual separation, time-consuming and complex identification as well as separation. Along with the use of the latest technologies (the YOLOv8 algorithm with adding the attention mechanism, stereo matching technology to define the 3D positioning, and the D-H model that defines kinematics of robotic arms is used), the system can be very precise and efficient in the process of sorting the aggregates representing different sizes, shapes, and lithologies. Explanatory experimental data show the general success with an average of 97.5 percent grasp and classification indicating the strength of the system and its flexibility. Though it has been successful, issues like understanding small aggregates and the misclassification due to their texture are some of the issues which can be improved on in the future. The modular billing technology and the versatile algorithms of the system highlight its future potential to help increase the productivity and lower the costs of operation as well as guarantee the safety within a workplace in such fields as construction, mining, and recycling. This study helps develop smart automation in a manufacturing environment, which demonstrates the paradigm-shifting abilities of combining robotics, computer vision, and artificial intelligence. In future, one can work on improving the way in which the system has to deal with low level aggregates and make the target detection procedures even more optimized in terms of their accuracy and reliability. The results of this research can serve as the basis of creating fully autonomous sorting systems, thus launching more environmentally friendly and efficient industrial processes.

**Author contributions**
Md. Taherul Islam Shawon: Writing – original draft, Visualization, Validation, Methodology, Investigation, Formal analysis, Data curation, Conceptualization. Yincai CAI: Writing – review & editing, Methodology, Software. Yuan LI: Writing – review & editing, Project administration, Funding acquisition, Conceptualization. Junjie NIU: Writing – review & editing, Investigation, software. Ting PENG: Writing – review & editing, Validation, Supervision, Project administration, Methodology, Investigation, Funding acquisition, Data curation, Conceptualization.

**Supporting information**
This work was supported by the National Natural Sciences Foundation of China under Grant 52378430.

**Declaration of Competing Interest**

The authors declare no potential conflicts of interests.

**References**

1. Song, R., et al., *A robotic automatic assembly system based on vision.* Applied Sciences, 2020. **10**(3): p. 1157.
2. Liu, J., et al., *Service platform for robotic disassembly planning in remanufacturing.* Journal of Manufacturing Systems, 2020. **57**: p. 338-356.
3. Jiang, T., et al., *A measurement method for robot peg-in-hole prealignment based on combined two-level visual sensors.* IEEE Transactions on Instrumentation and Measurement, 2020. **70**: p. 1-12.
4. Zhao, Q., et al., *Monocular vision-based parameter estimation for mobile robotic painting.* IEEE Transactions on instrumentation and measurement, 2018. **68**(10): p. 3589-3599.
5. Li, F., et al., *Manipulation skill acquisition for robotic assembly based on multi-modal information description.* IEEE Access, 2019. **8**: p. 6282-6294.
6. Hanh, L.D. and V.D. Cong, *Implement contour following task of objects with unknown geometric models by using combination of two visual servoing techniques.* International Journal of Computational Vision and Robotics, 2022. **12**(5): p. 464-486.
7. Cai, C., et al., *A novel measurement system based on binocular fisheye vision and its application in dynamic environment.* IEEE Access, 2019. **7**: p. 156443-156451.
8. Cong, V.D. and D.A. Duy, *Design and Development of Robot Arm System for Classification and Sorting Using Machine Vision.* FME Transactions, 2022. **50**(1).
9. El-Shair, Z.A. and S.A. Rawashdeh, *Design of an object sorting system using a vision-guided robotic arm.* 2019.
10. Vo, C.D., D.A. Dang, and P.H. Le, *Development of multi-robotic arm system for sorting system using computer vision.* Journal of Robotics and Control (JRC), 2022. **3**(5): p. 690-698.
11. Wang, Z., H. Li, and X. Zhang, *Construction waste recycling robot for nails and screws: Computer vision technology and neural network approach.* Automation in Construction, 2019. **97**: p. 220-228.
12. Chen, S., Y. Li, and N.M. Kwok, *Active vision in robotic systems: A survey of recent developments.* The International Journal of Robotics Research, 2011. **30**(11): p. 1343-1377.
13. Pande, B., et al. *A review of image annotation tools for object detection.* in *2022 International Conference on Applied Artificial Intelligence and Computing (ICAAIC).* 2022. IEEE.
14. Freeman, H. and R. Shapira, *Determining the minimum-area encasing rectangle for an arbitrary closed curve.* Communications of the ACM, 1975. **18**(7): p. 409-413.
15. Suzuki, S., *Topological structural analysis of digitized binary images by border following.* Computer vision, graphics, and image processing, 1985. **30**(1): p. 32-46.
16. Agarwal, R.P. and H. Agarwal, *Origin of irrational numbers and their approximations.* Computation, 2021. **9**(3): p. 29.
17. Blott, S.J. and K. Pye, *Particle shape: a review and new methods of characterization and classification.* Sedimentology, 2008. **55**(1): p. 31-63.
18. Ulusoy, U., *A review of particle shape effects on material properties for various engineering applications: from macro to nanoscale.* Minerals, 2023. **13**(1): p. 91.
19. Petersen, R., *Ubuntu 24.04 LTS Server: Administration and Reference.* 2025: surfing turtle press.
20. Ardana, F., I. Kanedi, and E. Prasetyo, *Time Limit Based Wireless Network Design Using Linux Ubuntu Server 20.04.* Jurnal Komputer Indonesia, 2023. **2**(2): p. 67–78-67–78.
21. Van Rossum, G. *Python programming language.* in *USENIX annual technical conference.* 2007. Santa Clara, CA.
22. Himagirish, T.S., et al., *Enhancing Real-Time Object Detection in Robotics through 3D Vision Integration.*
23. BELAHOUEL, R., *Machine Learning for Solving Inverse Kinematics of a 5-DOF Robotic Arm Case Study: LeArm by Hiwonder Lewan Soul.* 2023.
24. Ruppert, N. and K. George. *Robotic Arm with Obstacle Detection Designed for Assistive Applications.* in *2022 IEEE World Conference on Applied Intelligence and Computing (AIC).* 2022. IEEE.
25. Liang, D., et al. *The Battery Management System with DNN Deployed in STM32.* in *2024 China Automation Congress (CAC).* 2024. IEEE.
26. Żyliński, M., A. Nassibi, and D.P. Mandic, *Design and implementation of an atrial fibrillation detection algorithm on the ARM cortex-M4 microcontroller.* Sensors, 2023. **23**(17): p. 7521.
27. Reshma, K. and R. Sreenath. *Design and implementation of an isolated switched-mode power supply for led application.* in *2016 International Conference on Computation of Power, Energy Information and Commuincation (ICCPEIC).* 2016. IEEE.
28. Otoo, L., *Hybrid Charging Station for Autonomous Robot Using Real Time Kinematics.* 2024.
29. Lou, H., et al., *DC-YOLOv8: small-size object detection algorithm based on camera sensor.* Electronics, 2023. **12**(10): p. 2323.
30. Yuan, C., et al., *Lithology identification by adaptive feature aggregation under scarce labels.* Journal of Petroleum Science and Engineering, 2022. **215**: p. 110540.
31. Bustillo Revuelta, M., *Aggregates*, in *Construction Materials: Geology, Production and Applications.* 2021, Springer. p. 17-53.
32. Park, J., R. Delgado, and B.W. Choi, *Real-time characteristics of ROS 2.0 in multiagent robot systems: an empirical study.* IEEE Access, 2020. **8**: p. 154637-154651.
33. Tsado, T.Y., *A comparative analysis of concrete strength using igneous, sedimentary and metamorphic rocks (crushed granite, limestone and marble stone) as coarse aggregate.* Zeszyty Naukowe Politechniki Częstochowskiej. Budownictwo, 2013.
34. Wang, K., Y. Hou, and C. Chen. *Influence of the earth rotation on trajectory of a returnable hypersonic cruise vehicle.* in *IOP Conference Series: Materials Science and Engineering.* 2018. IOP Publishing.
35. Sun, J.-D., et al. *Analytical inverse kinematic solution using the DH method for a 6-DOF robot.* in *2017 14th international conference on ubiquitous robots and ambient intelligence (URAI).* 2017. IEEE.
36. Lu, P., Q. Liu, and J. Guo. *Camera calibration implementation based on Zhang Zhengyou plane method.* in *Proceedings of the 2015 Chinese Intelligent Systems Conference: Volume 1.* 2015. Springer.
37. Zhang, Z., *Camera calibration with one-dimensional objects.* IEEE transactions on pattern analysis and machine intelligence, 2004. **26**(7): p. 892-899.
38. Fetić, A., D. Jurić, and D. Osmanković. *The procedure of a camera calibration using Camera Calibration Toolbox for MATLAB.* in *2012 Proceedings of the 35th International Convention MIPRO.* 2012. IEEE.
39. Kang, S., S.D. Kim, and M. Kim, *Structural-information-based robust corner point extraction for camera calibration under lens distortions and compression artifacts.* IEEE Access, 2021. **9**: p. 151037-151048.
40. Atkinson, D. and T.H. Becker, *Stereo digital image correlation in MATLAB.* Applied Sciences, 2021. **11**(11): p. 4904.
41. Zhang, Z., *A flexible new technique for camera calibration.* IEEE Transactions on pattern analysis and machine intelligence, 2002. **22**(11): p. 1330-1334.
42. Shiu, Y.C. and S. Ahmad, *Calibration of wrist-mounted robotic sensors by solving homogeneous transform equations of the form AX= XB.* 1987.
43. Tsai, R.Y. and R.K. Lenz, *A new technique for fully autonomous and efficient 3 d robotics hand/eye calibration.* IEEE Transactions on robotics and automation, 1989. **5**(3): p. 345-358.
44. Chesher, C. and F. Andreallo, *Eye machines: Robot eye, vision and gaze.* International journal of social robotics, 2022. **14**(10): p. 2071-2081.